\documentclass[conference]{IEEEtran}
\usepackage{cite}
\usepackage{amsmath,amssymb,amsfonts}
\usepackage{algorithmic}
\usepackage{graphicx}
\usepackage{textcomp}
\usepackage{xcolor}
\usepackage{qtree}
\def\BibTeX{{\rm B\kern-.05em{\sc i\kern-.025em b}\kern-.08em
    T\kern-.1667em\lower.7ex\hbox{E}\kern-.125emX}}
\begin{document}

\title{Understanding Deep Learning using Topological Dynamical Systems, Index Theory, and Homology}

\author{\IEEEauthorblockN{Bill Basener}
\IEEEauthorblockA{\textit{Professor of Data Science} \\
\textit{University of Virginia, School of Data Science}\\
Charlottesville, VA \\
wb8by@virginia.edu}
}

\maketitle

\begin{abstract}
In this paper we investigate Deep Learning Models using topological dynamical systems, index theory, and computational homology. These mathematical machinery was invented initially by Henri Poincar\'e around 1900 and developed over time to understand shapes and dynamical systems whose structure and behavior is too complicated to solve for analytically but can be understood via global relationships.  In particular, we show how individual neurons in a neural network can correspond to simplexes in a simplicial  complex manifold approximation to the decision surface learned by the NN, and how these simplexes can be used to compute topological invariants from algebraic topology for the decision manifold with an explicit computation of homology groups by hand in a simple case.  We also show how the gradient of the probability density function learned by the NN creates a dynamical system, which can be analyzed by a myriad of topological tools such as Conley Index Theory, Morse Theory, and Stable Manifolds.  We solve analytically for associated the differential equation for a trained NN with a single hidden layer of 256 Neurons applied to the MINST digit dataset, and approximately numerically that it a sink and basin of attraction for each of the 10 classes, but the sinks and strong attracting manifolds lie in regions not corresponding to images of actual digits.  Index theory implies the existence of saddles.  Level sets of the probability functions are 783-dimensional manifolds which can only change topology at critical points of the dynamical system, and these changes in topology can be investigated with Morse Theory.
\end{abstract}


\section{Introduction}
\label{introduction}
Topology is the study of shape apart from angles and distance.  For example, a triangle is topologically equivalent to a circle, but distinct from a figure-eight.  Combinatorial topology is the study of combinations of simple pieces; for example collections of line segments connected at endpoints, surfaces constructed by connecting triangles connected along their edges, or higher dimensional manifold created be connecting tetrahedra or higher dimensional analogs called simplexes along their faces.  In a similar way, neural networks are constructed of simple components (neurons), and it is the global properties between the neurons from the weights that define the profound learned information.  In this paper we provide computation of the set separating classes learned by a neural network as a set of line segments connected at their endpoints, and explicitly compute a topological invariant for this set called the homology groups.  The example is simple for the sake of exposition, but the general principle that weights and architecture of a neural network can be used to compute topological invariants is very interesting.

An important use of topology that motivated its development is the ability to understand behavior in dynamical systems (for example, flows or solutions to differential equations of a vector field) even when the computation of analytic solutions unreasonably complicated or impossible.  There is a rich variety of mathematical machinery that has been developed along this generally calculus-based approach.  Morse theory describes the topology of a manifold based on a gradient function, and index theory gives relationships between the topology of manifolds and vector fields.  For example the 'hairy ball' theorem says that any vector field on a 2-dimensional sphere with have locations where the vector length is zero, and in addition the sum of the indexes for these locations will equal the Euler characteristic for the sphere, which is 2.  The hairy ball theorem is an example of the Poincaré–Hopf index theorem (the sum of the indexes of a vector field on a manifold equals the Euler characteristic of the manifold), and more generally Conley Index Theory and The Fundamental Theorem of Dynamical Systems~\cite{norton1995fundamental}.

In this paper we consider the vector field on the input space $X$ for a neural network defined by the gradient of the probability function for a class $G$, specifically $\nabla P_G(x)$ for $x\in X$ and where $P_G(x)$ is the probability for class $G$ assigned to input $x$.  One can also study the gradient field $\nabla P(x)$ where $P(x)=\textrm{max}\{P_G(x)\}$ where the max is over all classes $G$; this vectorfield is defined everywhere except the separating manifolds.  We construct a neural network trained on the standard MINST dataset, explicitly compute the gradient vectorfield from the weights in the network, and follow the orbits of the MINST data under the flow of this gradient vector field.  It appears there are ten sinks (maxima for the gradient), and the basic of attraction for each is the classification region for each class.  Interestingly, digit image for every sink has no resemblance to the digit it is representing, an interesting connection to adversarial networks.  We suggest that, in general, a neural network learns weights, but at the same time what is being learned is probability density functions on the (often high-dimensional) input space, and the topology of these pdfs can be studied by looking at the gradient functions, with a rich mathematical theory being available for connecting topology with vector fields.

Topological methods have been used to study neural networks recently.  For example, it has been shown that viewing the data passung through successive layers, the topology of the classes tends to become simpler from each layer to the next through decreasing Betti numbers, computational homology, and visual inspection~\cite{naitzat2020topology}. Neural networks have been used to learn topological signatures or features in images, with layers to exploit these features for learning tasks~\cite{hofer2017deep}.  Manifold geometry and topology has been used to study manifolds that approximate data clouds~\cite{hofer2017deep}.  In this paper, our emphasis is on using topology and dynamical systems theory to understand the classification structures, separating manifolds, and probability density functions learned by a network and used for further prediction.

While beyond the scope of this paper, it is worth noting that our two approaches, combinatorial topology from discrete pieces of a network/manidolf and vectorfields with calculus, are connected in a deeply mathematical sense.  The homology groups we compute are computed from combinations of simplicies.  There are theories of cohomology groups defined using integrals of vectors fields and more generally differential forms on manifolds.  De Rham's Theorem says that under certain definitions for these groups and assumptions, the discrete-based homology groups are isomorphic to the calculus-based chomology groups.  Hopefully this gives some sense of coherence between our two topological approaches to study 'what is learned' by a neural network.

\section{Homology of Separating Manifold: A Simple Example}
\label{homology}
\begin{figure}[ht]
\vskip 0.2in
\begin{center}
\centerline{\includegraphics[width=\columnwidth]{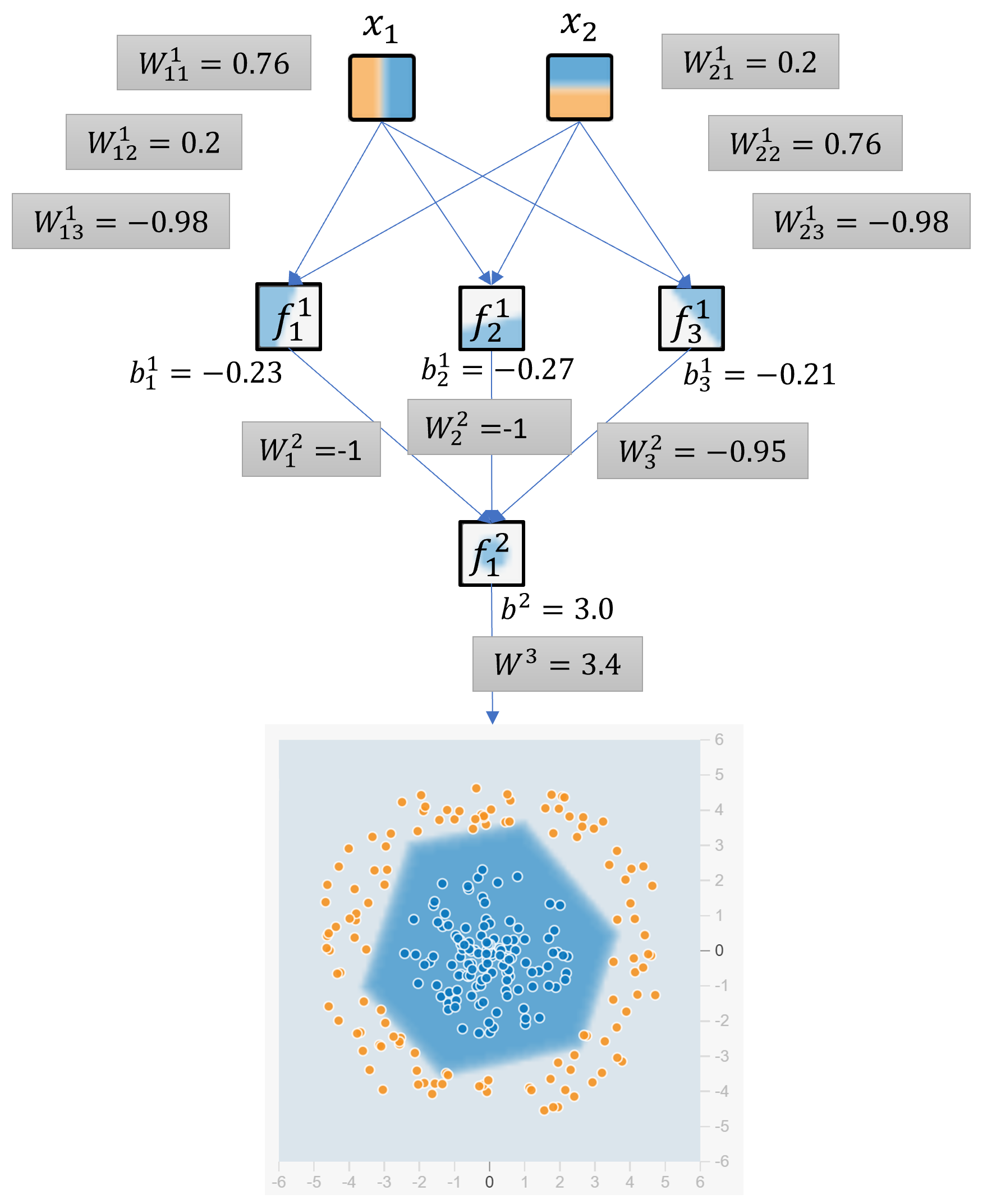}}
\caption{A simple neural network with trained weights.}
\label{simpleNN}
\end{center}
\vskip -0.2in
\end{figure}

We constructed the neural netowork with two classes, a single hidden layer of 3 ReLU neurons, and a final classification layer with a single neuron for classification.  The network with trained weights on a simple 2-class problem is shown in Figure~\ref{simpleNN}.

\begin{figure}[ht]
\vskip 0.2in
\begin{center}
\centerline{\includegraphics[width=\columnwidth]{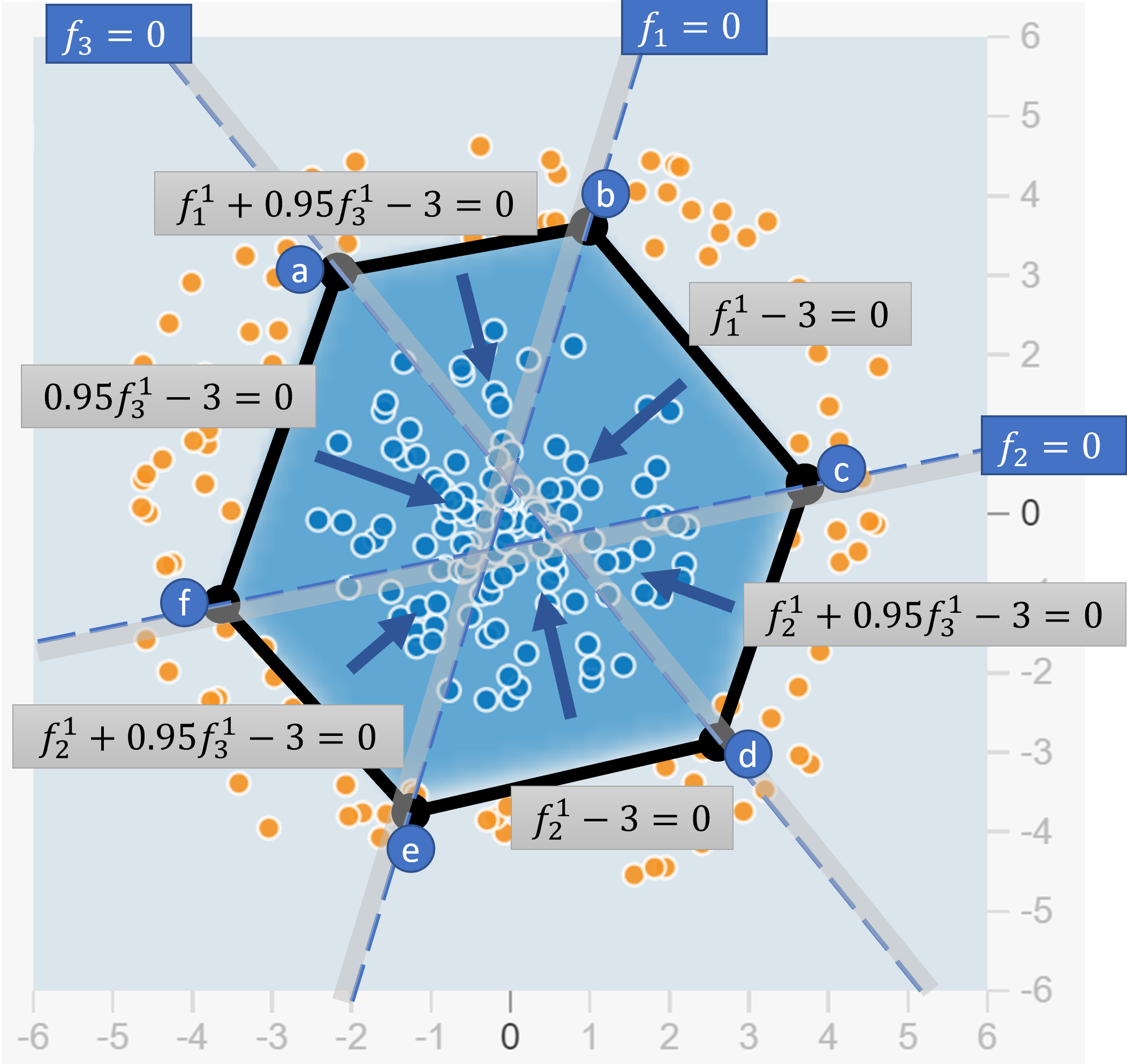}}
\caption{The separating surface where the class probabilities are equal is a set of line segments which make up the hexegon.  The vertices of the hexagon are labeled $a,b,c,d,e,f$.  The output of the neural network is positive inside the hexagon and zero outside.  For each $f_i^1$, the line where $f_i^1=0$ is shown as a bold dashed line, and the shaded side indicates the side where the output of the ReLU $F_i^1$ is increasing.  The arrows show the direction of increase in output of the neural network.  Inside the center triangle, $f_1=f_2=f_3=0$ and this the output of the network is $b_2*W^2=3\times3.4=11.4$.}
\label{boundaryHomology}
\end{center}
\vskip -0.2in
\end{figure}

It is not hard then to show that the boundary between the classes is the set $f_1^2(x_1,x_2)=0$ consists of the line segments labeled with their formulas as shown in Figure~\ref{boundaryHomology}, where $f_i^1(x_1,x_2)$ is the output of the ReLU function with the inputs and biases shown in figure~\ref{simpleNN}.

For a definition of homology and background material, see~\cite{Hatcher2000}, but some definitions are as follows.  An $n-$dimensional simplex (or just \textit{$n-$simplex}) $(n\geq 0)$ is the convex hull of $n+1$ points (called vertices) such that no three of the vertices are coplanar; for example a point, a line segment, a triangle, tetrahedron, etc.  A \textit{simplicial complex} is a set of simplexes such that if any two simplexes intersect, they do so along one faces (the lower dimensional sub-simplexes).  The separating set $S=\{(x_1,x_2)|f_1^2(x_1,x_2)=0\}$ is a simplicial complex consisting of six 1-dimensional simplexes and six 0-dimensional simplexes (points).  It is a continuous (but not smooth) 1-dimensional manifold.

For a simplicial complex, the group of $n-$chains is
\[
C_n(X)=\{\sum_i m_i \sigma_i |m_i\in\mathbb{Z}\}
\]
and each $\sigma_i$ is an $n-$dimensional simplex in the complex.  The boundary $\partial$ of an $n-$simplex is the set of $(n-1)-$dimensional faces (any convex hull of $(n-1)$ of the vertices, which is of course itself an $(n-1)$-dimensional simplex) with the orientation inherited from the simplex.  Then $\partial_k:C_k(X)\to C_{k-1}(X)$, and the $k^\textrm{th}$ homology group of $X$ is defined to be
\[
H_k(X)=\text{ker}(\partial_k)/\textrm{Im}(\partial_{k+1}).
\]
Clearly, for the separating manifold $S$,
\[
H_n(S)=0,\textrm{ for }n>1
\]
since there are no $n-$chains with $n>1$.  The group $C_1$ has the basis $\{\overline{ab},bc,cd,de,ef,fa\}$ and $C_0$ has the basis $\{a,b,c,d,e,f\}$.  Clearly $\partial_2(mmab+mbc+mcd+mde+mef+mfa)=0$ for any integer $m$ and this defines $\textrm{ker}(\partial_1)$.  Then
\[
H_1(S)= \text{ker}(\partial_1)/\textrm{Im}(\partial_2 \cong \mathbb{Z}/0 \cong \mathbb{Z}
\]
and
\[
H_1(S)= C_0/\textrm{Im}(\partial_1) \cong \mathbb{Z}.
\]
This shows that the separating manifold (curve) has the homology groups of a circle, and thus is (isomorphic to) a circle.  While this example is trivial, it suggests the tools of computational topology (an excellent text is~\cite{Edelsbrunner2010} while introductory topics are covered in~\cite{Basener2007} and also see~\cite{Kaczynski2004}) may be useful in determining homology for more complex situations.  At minimum, it shows that the homology groups can be computed from just knowing the weights and architecture.

\section{Dynamical Systems and Index Theory}
\begin{figure}[ht]
\vskip 0.2in
\begin{center}
\centerline{\includegraphics[width=\columnwidth]{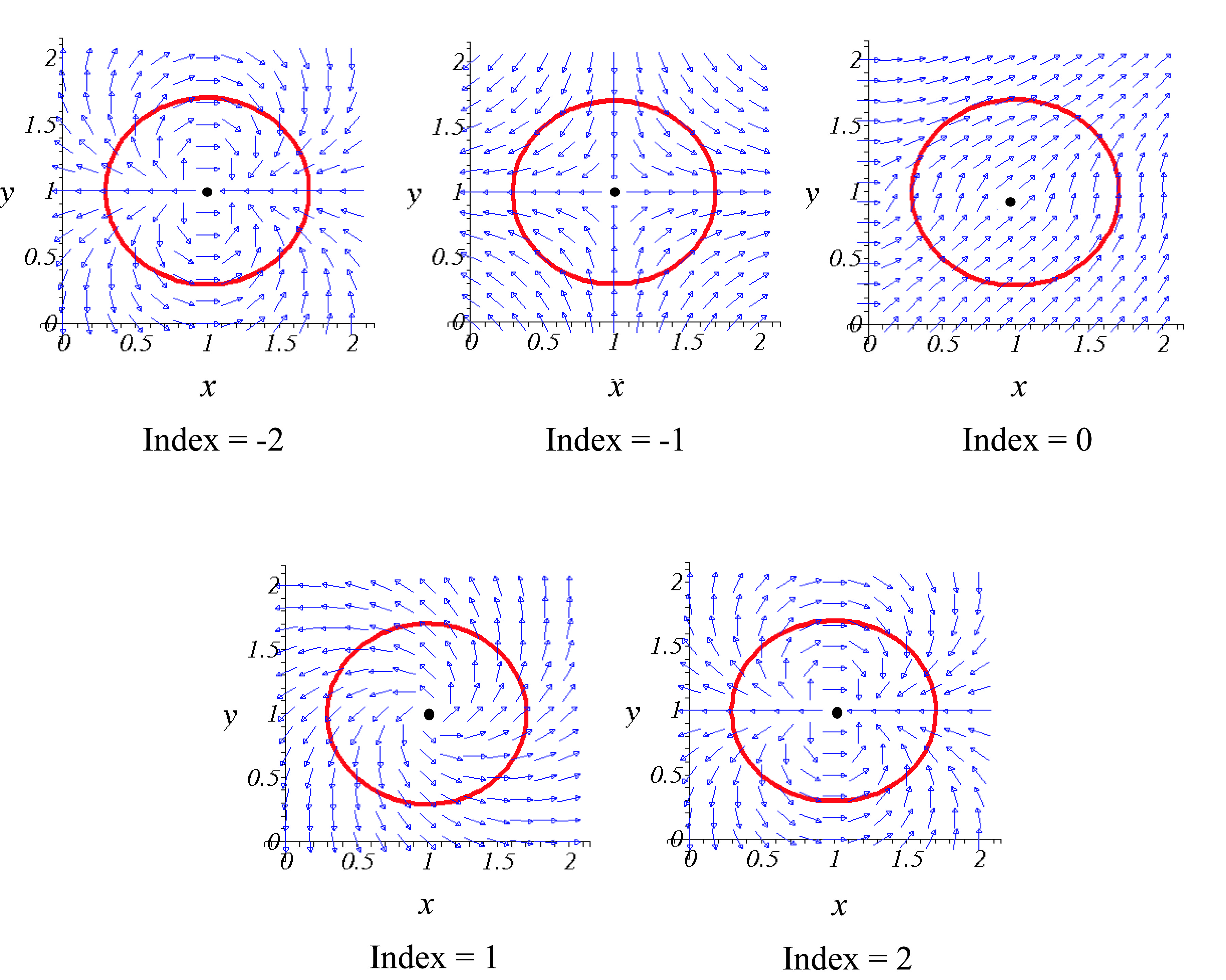}}
\caption{The index for a few different vector fields in 2-dimensions, in which case this index is equal to the number of counterclockwise rotations of the vector as you travel counterclockwise around a circle isolating the fixed point.  (Figure is from~\cite{Basener2007} used with permission.)}
\label{Index}
\end{center}
\vskip -0.2in
\end{figure}

A central theorem in dynamical systems and index theory is the Poincar\'e-Hopf Index theorem, which says that for a vectorfield $v:M\to\mathbb{R}^n$ on an $n-$dimensional manifold $M$, the sum of the index of the fixed points (aka zeros, equilibria, or singularities) is equal to the Euler characteristic of the manifold,
\[
\sum_i I(x_i) = \chi (M),
\]
where the index $I$ of a fixed point (or zero) of the vector field is defined by taking an isolating sphere $S$ around the fixed point and the index is equal to the degree of the map $S \to S$ defined by
\[
x\mapsto v(x)/||V(x)||.
\]
Poincar\'e proved this in two dimensions, and Hopf extended it to arbitrary dimensions.  Examples are shown in Figure~\ref{Index}, which is in two dimensions where the index is equal to the number of counterclockwise rotations of the vector as you travel counterclockwise around a circle isolating the fixed point.
There are numerous good sources on the Index Theory for vectorfields and applications such as~\cite{zhang2006vector, josevichpoincare, reininghaus2011combinatorial, libgober2012euler} and work on computing and visualizing the topology of vector fields~\cite{Scheuermann1998, Reininghaus2011}.  This theory was extended by Cayley and James Clerk Maxwell to study topography, and generalized to gradients of functions on manifolds by Marsten Morse in what is now called Morse Thoery~\cite{milnor2016morse}.  This line of theory culminated in Conley Index theory published by Conley in~\cite{conley1978isolated} focusing on isolated invariant sets (not just fixed points), and also see~\cite{mischaikow2002conley}, and is connected to Floer homology~\cite{salamon1990morse}.  One of Conley's theorems proving all dynamical systems are composed of (chain recurrent) invariant sets and gradient-like sets known as the The Fundamental Theorem of Dynamical Systems~\cite{norton1995fundamental}.  The triangle in the center of Figure~\ref{boundaryHomology} is an isolated invariance set under the flow moving in the direction of increasing probability.  A deep and profound concept throughout these topics, evident even in the 2-dimensional vector fields, is that discrete objects like the index of a fixed point or Euler characteristic of a manifold, which are stable under continuous perturbations, are connected to integrals and differential constructions.

\begin{figure}[ht]
\vskip 0.2in
\begin{center}
\centerline{\includegraphics[width=\columnwidth]{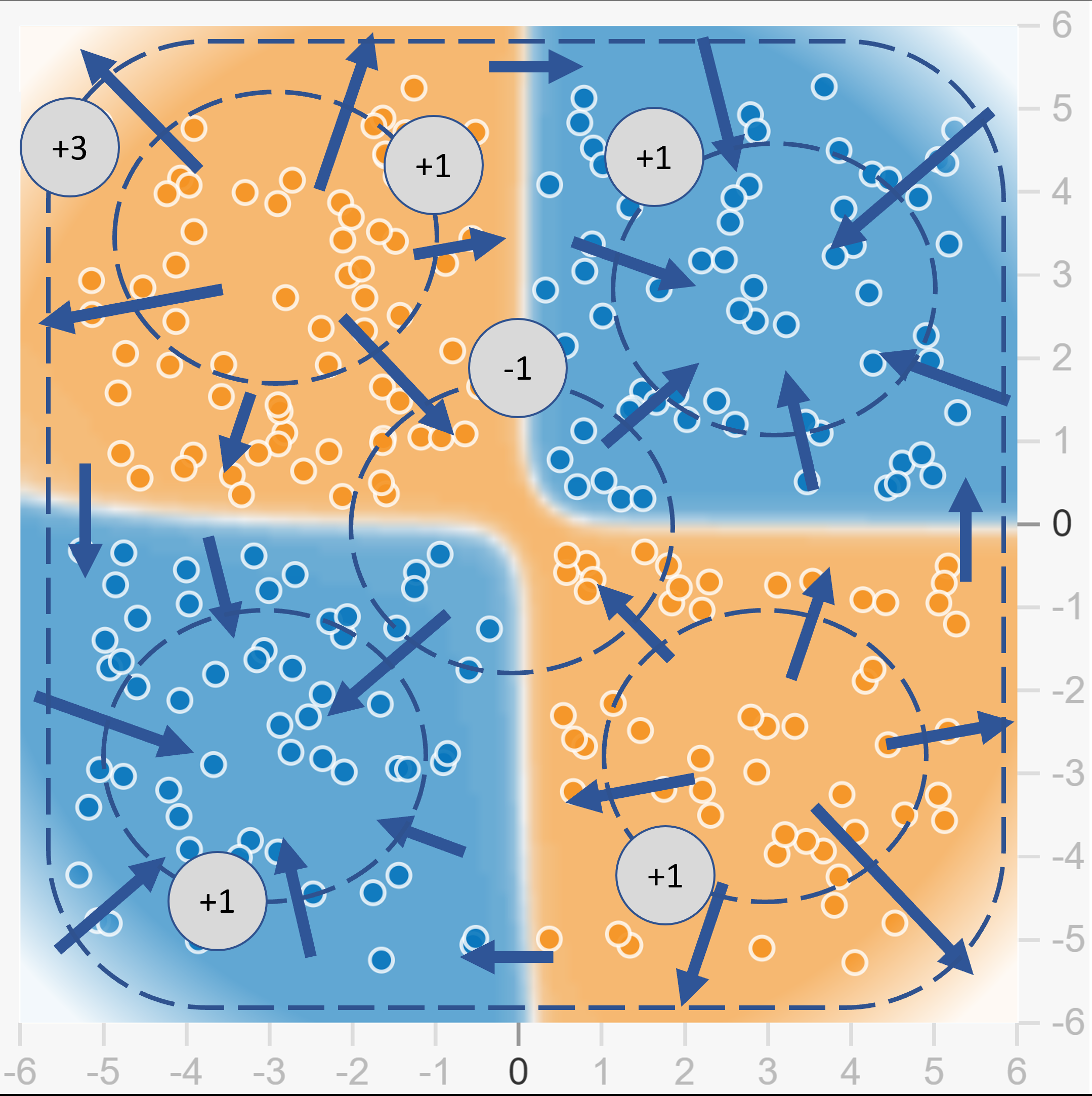}}
\caption{Computations of the degrees of isolated invariant sets by counting rotations of the vector field around a isolating curve (topological circle), where the vectorfield is the gradient of the probability output of the NN for the class shown in blue.  Observe that the index around the outer cueve equals the sum of the indices of the smaller circles.  This is the Poincar\'e Hopf Theorem applied to the probability gradient of a NN.  (This is a notional depiction of a neural network for the sake of illustrating the theorem created by taking the output of a NN and modifying so the probabilities close to zero around the edges of the region $[-6,6]\times[-6,6]$ and arrows are added visually to illustrate the theorem.)}
\label{PHThmDL}
\end{center}
\vskip -0.2in
\end{figure}

We suggest that the concepts of index theory and related topological tools can be used to study the structure of 'what is learned' by a neural network.  Specifically, we can apply index theory and general theory for dynamical systems to the gradient vector field of probability functions on the state space that are learned by the network.  We do this by training a neural network on the MINST dataset and solving explicitly for the gradient of the probability functions for the classes.  The input images are $28$ by $28$ pixels (so our vector field is in $28\times28=784$-dimensional space) and we have a single hidden layer of 256 ReLU neurons and an output layer of 10 softmax classification neurons (one for each of the 0 through 9 digit classes).  The network has a $97\%$ validation accuracy which is reasonable but relatively well below what can be achieved by deep CNNS which can achieve $99.77\%$ accuracy or better~\cite{ciregan2012multi}.

Our input space is $\mathbb{R}^784$ (restricted the unti cube with all values in $[0,1]$), and we denote the weight from input $i$ to neuron $j$ in the first layer by $W_{ij}^1$, with these weights making the $784\times 256$ matrix $W^1$.  The weight from hidden layer neuron $i$ to softmax neuron $k$ is denoted $W_{jk}^2$, and these weights make up the $256\times 10$ matrix $W^2$.  The biases are denoted by $b_j^1$ for the $j$-th neuron in the first layer and $b_k^2$ for the $k$-th neuron in the classification layer.  We denote the $j$-th neuron in the hidden layer by $f_j$ and the $k$-th neuron function in the softmax classification layer by $g_k$.  The gradient is a simple application of the chain rule, with
\[
\nabla g_k(x_1,..,x_784) = \left( \frac{\partial g_k}{\partial x_1},...,\frac{\partial g_k}{\partial x_{784}}\right)
\]
where
\[
\frac{\partial g_k}{\partial x_i} = \sum_{j=1}^{256} \frac{\partial g_k}{\partial f_j} \frac{\partial f_j}{\partial x_i}.
\]

\begin{figure}[ht]
\vskip 0.2in
\begin{center}
\centerline{\includegraphics[width=\columnwidth]{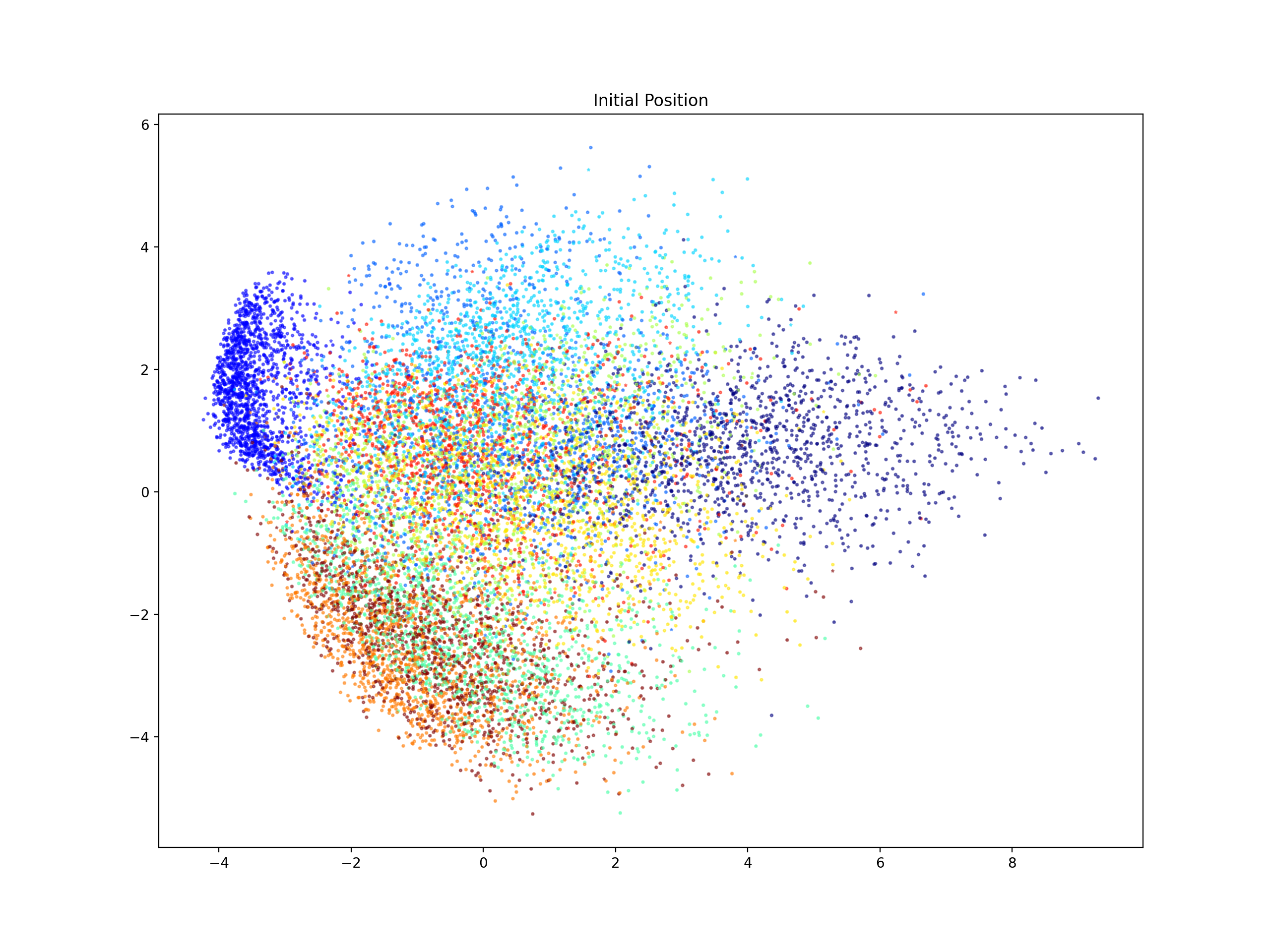}}
\caption{The MIST data shown with points colored by class membership.  This data is shown projected onto its first two PCs.}
\label{MINSTpca}
\end{center}
\vskip -0.2in
\end{figure}

\begin{figure}[ht]
\vskip 0.2in
\begin{center}
\centerline{\includegraphics[width=\columnwidth]{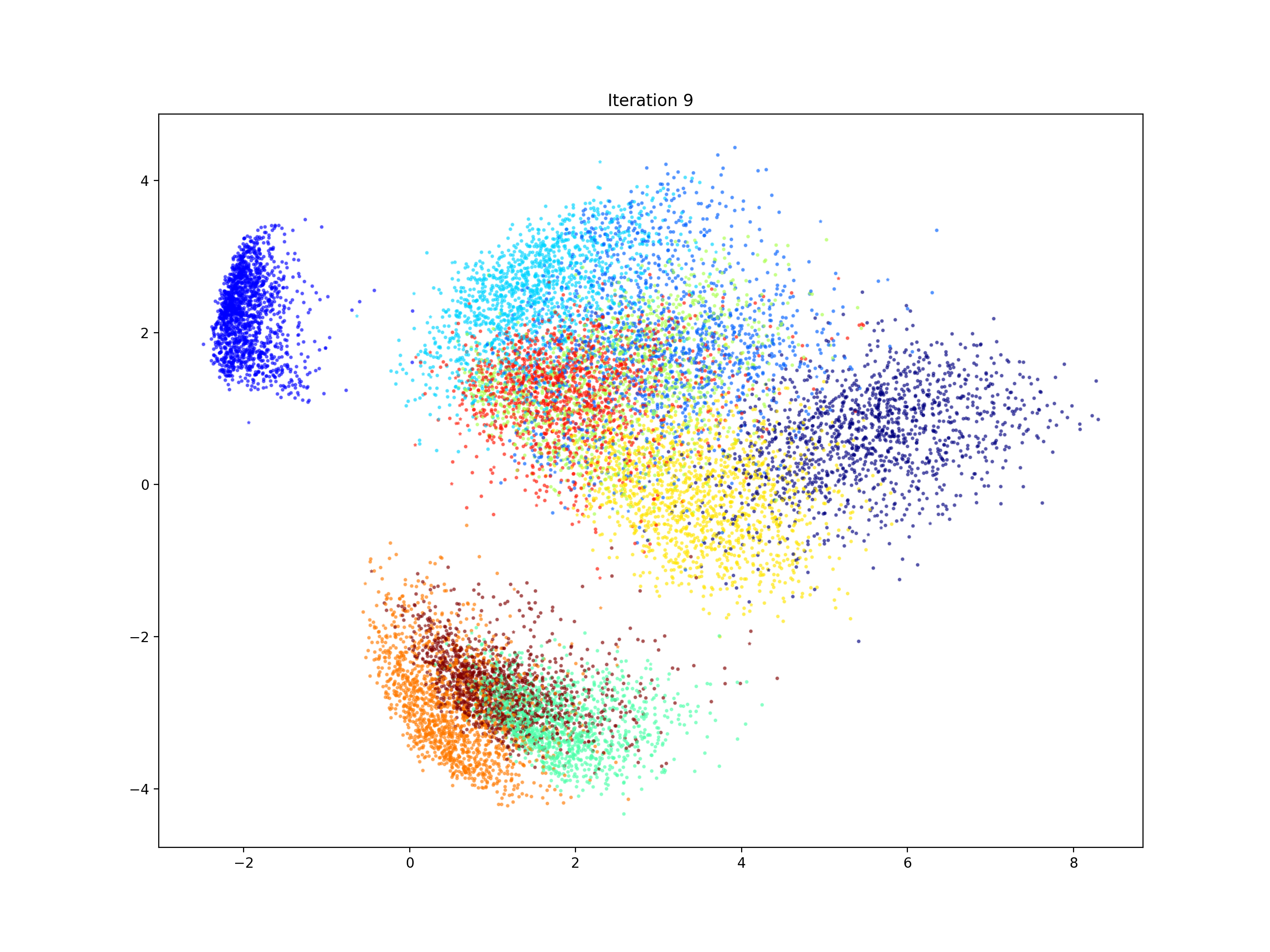}}
\caption{The MIST data after 9 iterations of the gradient flow Euler's approximation.  This data is shown projected onto the first two PCs computed from the original data.}
\label{Iteration_9}
\end{center}
\vskip -0.2in
\end{figure}

\begin{figure}[ht]
\vskip 0.2in
\begin{center}
\centerline{\includegraphics[width=\columnwidth]{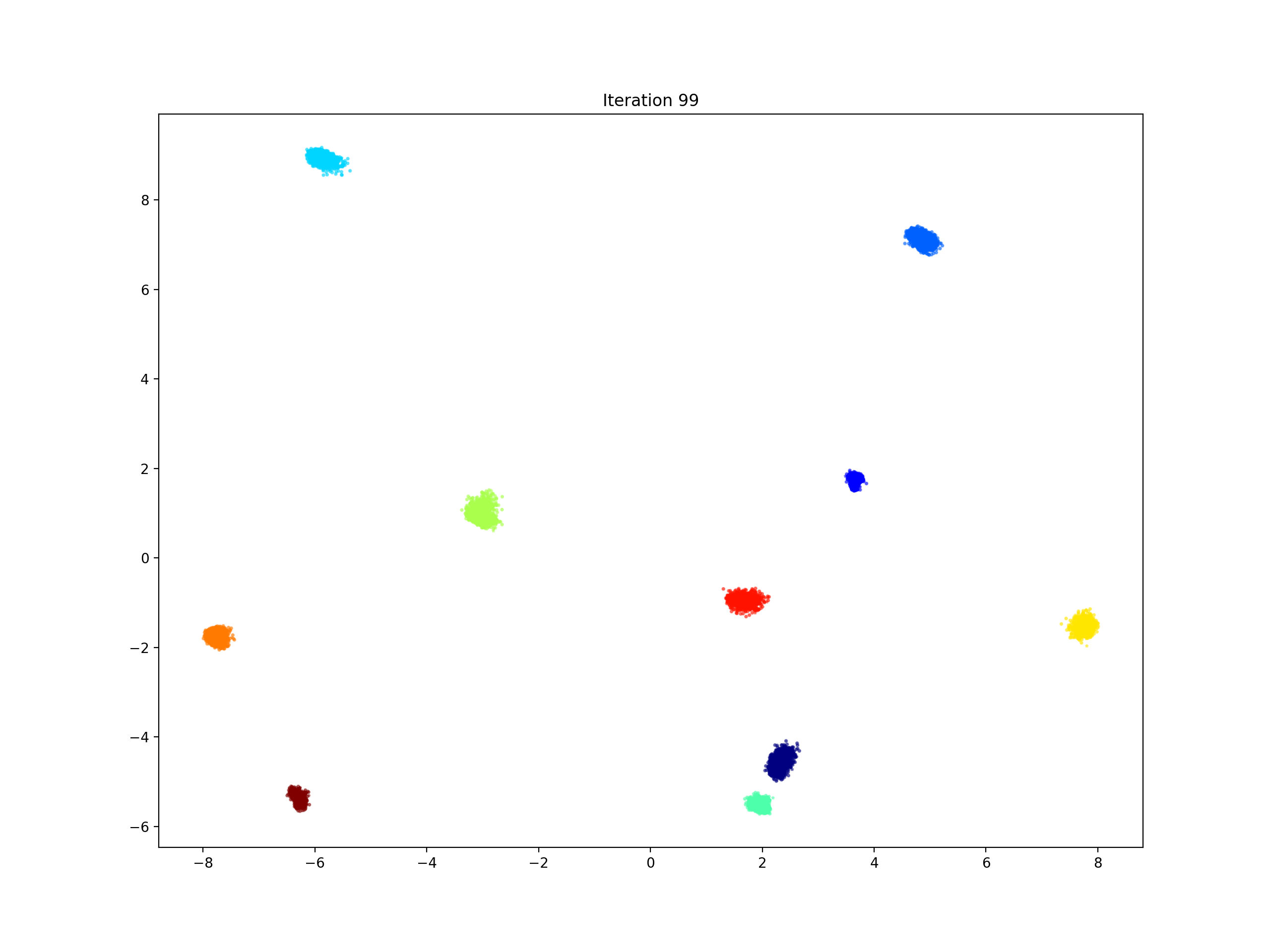}}
\caption{The MIST data after 99 iterations of the gradient flow Euler's approximation.  This data is shown projected onto its first to PCs.}
\label{Iteration_99}
\end{center}
\vskip -0.2in
\end{figure}

In Figure~\ref{MINSTpca} we show the projection of the MINST data on the first two PCAs colored by class.  Figure~\ref{Iteration_9} shows this data on the same PCA projection after 9 iterations of Eulers method approximation to the gradient flow $x'=\nabla g_k(x)$ stepping by $x \mapsto x + 0.05\nabla g_k(x)$. Figure~\ref{Iteration_99} show the data after 99 iterations, projected on the first two PCs computed from that data.

\begin{figure}[ht]
\vskip 0.2in
\begin{center}
\centerline{\includegraphics[width=\columnwidth]{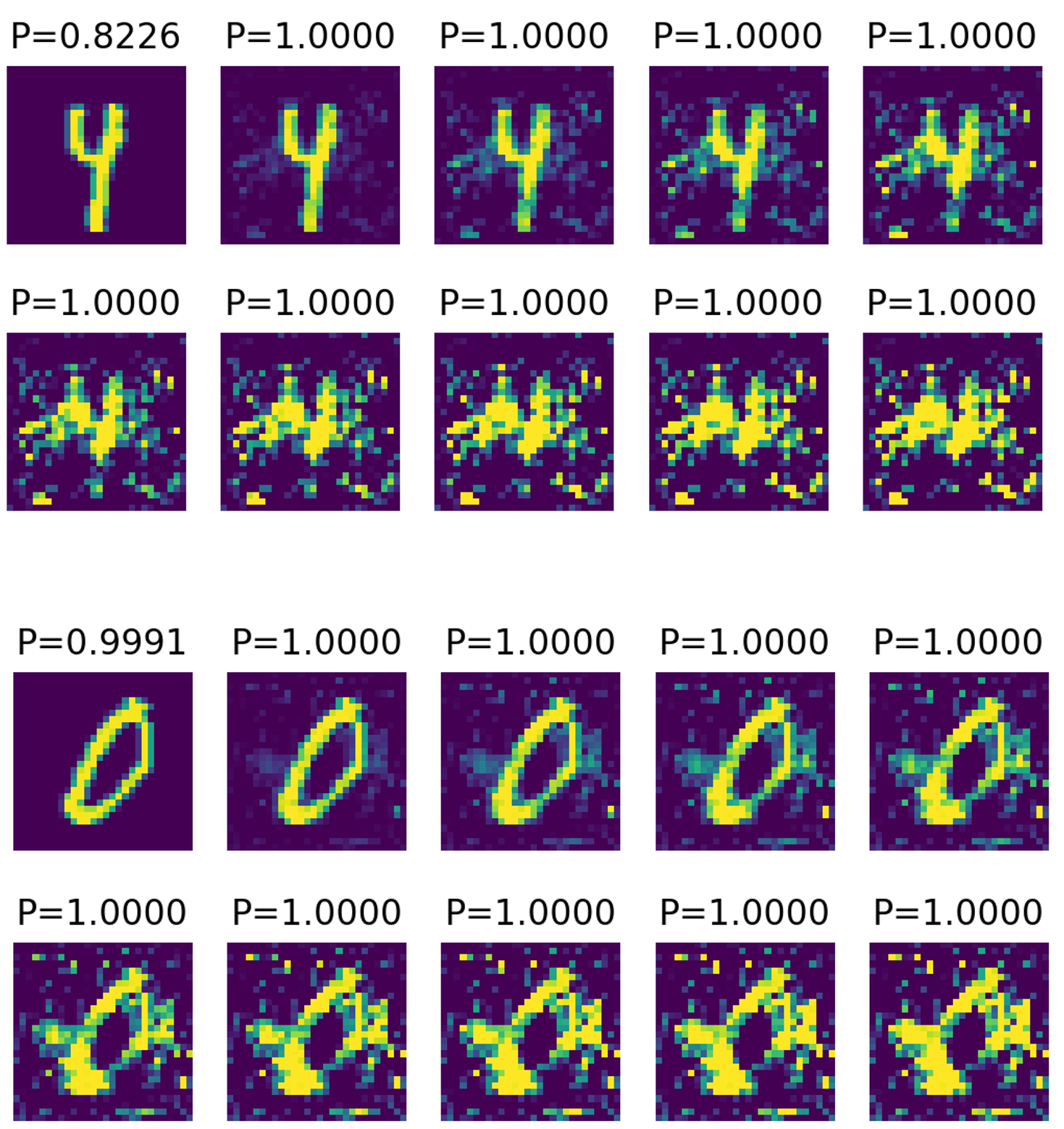}}
\caption{The first 10 iterations of a handwritten 4 digit is shown in the top 2 rows, and the first 10 iterations of a handwritten 0 digit is shown in the bottom 2 rows.  Note that with each successive iteration the probability increases but the image looks less like a number.}
\label{digitIterations}
\end{center}
\vskip -0.2in
\end{figure}

\begin{figure}[ht]
\vskip 0.2in
\begin{center}
\centerline{\includegraphics[width=\columnwidth]{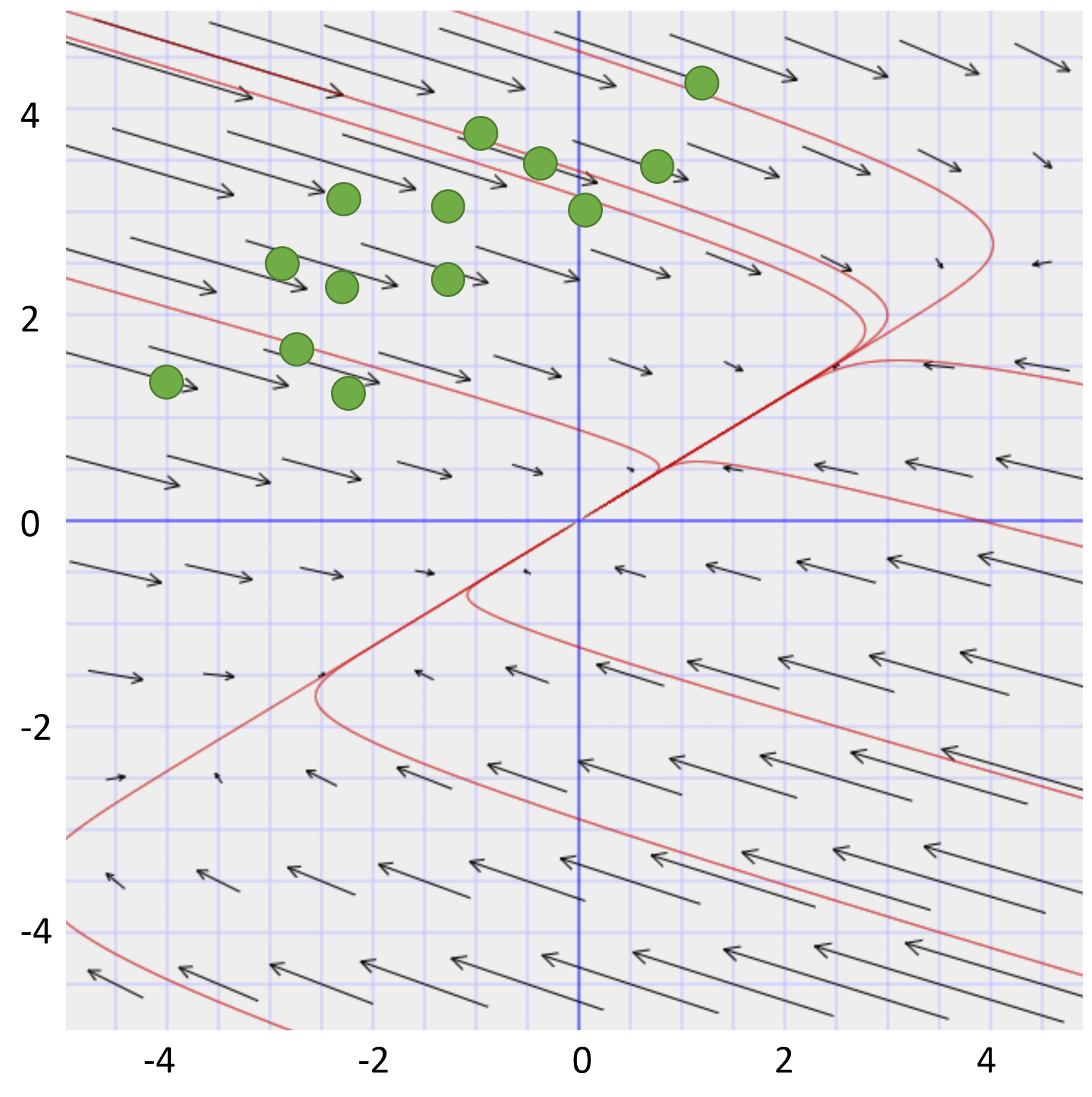}}
\caption{A vector field and phase plane for a 2D linear differential equation that has a strong attracting direction.  This is a notional depiction of what seems to be happening in the probability gradient flow for the NN on the MINST data, where the circles in the figure notionally represent a distribution of data that is off the strong attracting manifold, and the 1d strong attracting direction notionally represents a higher dimensional strong stable manifold.}
\label{notionalAttractingDirection}
\end{center}
\vskip -0.2in
\end{figure}

In Figure~\ref{digitIterations}, we shown the digit images for 10 iterations each of a 4 (top two rows) and a 0 (bottom two rows).  In each iteration the probability is increasing.  However, observe that the visual look of the image is becoming less like a digit.  This same phenomena is consistent across all digits observed.  Clearly, this is not desirable behaviour as the non-digit-like images are being classified with very high probability to a single class.  It seems there is a string attracting stable manifold for the probability max for each class; that is, there is a submanifold in the data space that is far from the actual data, and under the gradient flow orbits are attracted to this manifold and approach the max along this manifold.  The behavior is depicted notionally in Figre~\ref{notionalAttractingDirection} using the linear differential equation $x'=-4x+6y, y'=x-y2$.  While our network architecture is known to be sub-optimal, observed phenomena like this could be used to improve optimization strategies and cost functions.

The function $g_k(f_1,...,f_{256})$ is defined by
\[
g_k(f_1,...,f_{256}) = \frac{e^{\sum_j W_{jk}^2 f_{j}+b_k}}{ \sum_{k'}e^{\sum_j W_{jk'}^2 f_{j}+b_{k'}}}
\]
which leads to
\[
\frac{\partial g_k}{\partial f_j} = g_k \left[ W_{jk}^2 - \sum_{k'\neq k} W{jk'}^2 g_{k'} \right]
\]
Define $\delta_j(x)$ to be the function defined by
\[
\delta_j(x)=
\begin{cases}
      1                  & \text{if }\sum_i W_{ij}^2 x_i + b_j^1 > 0\\
      \textrm{undefined} & \text{if }\sum_i W_{ij}^2 x_i + b_j^1 = 0\\
      0                  & \text{otherwise}
\end{cases}
\]
(For numerical stability we use $\delta_j(x)=0$ when $\sum_i W_{ij}^2 x_i + b_j^1 > 0$ in our simulations.) The the partial derivatives of $f_j$ are defined by
\[
\nabla f_j = \left( \frac{\partial f_j}{\partial x_1},...,\frac{\partial f_j}{\partial x_{256}}\right) = \delta_j(x) (W_{1j}^1,...,W_{256j}^1)
\]
So
\[
\frac{\partial g_k}{\partial x_i} = \sum_{j=1}^{256} \delta_j(x) g_k \left[ W_{jk}^2 - \sum_{k'\neq k} W{jk'}^2 g_{k'} \right]W_{jk}^1
\]
More concisely
\[
\nabla g_k(x) = g_k(x) \widetilde{W}^1(x) \left[\Phi_k(x) + \Gamma(x) \right]
\]
where  $\Phi_k$ is the $256\times 1$ column vector whose $j$-th value is $(1+g_k(x))W_{jk}^2$; $\Gamma$ is the $256\times 1$ column vector $W^2g(x)$; and $\widetilde{W}^1(x)$ is a $784\times 256$ matrix
\[
\widetilde{W}^1_{ij}(x)=\delta_j(x)W_{ij}^1.
\]

We now consider the zeros and/or singularities for $\nabla g_k(x)$.  It is clear that $g_k(x)>0$ for all $x$.  The vector $\Phi_k$ is nonzero (except in the degenerate case where $W_{jk}^2=0$ for all $j=1,...,256$).  The vector $\Gamma$ is also nonzero.  Thus, the attracting sets (possibly simplexes) of the gradient flow are the points where $\widetilde{W}^1_{ij}(x)$ is undefined or zero.

\section{Conclusions}
In this paper we presented how the homology of a separating manifold can be computed from the weights in a simple neural network.  We suggest that this may be helpful in understanding the complexity of the separating surfaces/manifolds and complexity of learned information.  Our example used ReLU activation functions so the results were combinatorial and readily understood with combinatorial homology, but other activation functions would lead to smooth gradients and differential tolopogy tools that mirror the combinatorial ones (cohomology, Jacobians of equilibria points, etc.).

We then showed how dynamical systems and index theory can be applied to give insight to what is leaned by a neural network.  We showed an expository example in 2 dimensions where the NN and theorem can be understood and computed visually.  We then gave an example in 784 dimensions where numerically numerical approximations suggest that there are 10 attracting invariant sets.  (Notably there are 10 classes, so the gradient at each point is the gradient of the class probability for the most probable class.)  We showed that the analytic formula for the gradient can be computed explicitly for this example.  We also showed that this dynamical systems approach uncovers an apparent weakness in the neural network; although the accuracy is over 97\% on validation data, the attracting sets where each class probiablity is maximized correspond to images that look nothing line the original data of handwritten digits.

This work suggests a number of lines for future research.  Can topology be used to better understand the probability functions - for example are the basins of attraction for the attractors simply connected, in our case for MINST or under certain circumstances?  Examples shown in Figures~\ref{Index} and ~\ref{PHThmDL} show the obvious fact that basins of attraction are neither always connected nor always comprised of simply connected components.  The differential equations for the gradient flow are exact and the probability is a Lyapunov function, constraining the type of dynamics (no periodic orbits or chaos, for example).  The most general question seems to be what topological information can generally be computed for neural networks, and what are the connections between this topology and the weights and architecture of the network.

Computing topology is interesting, but is not only an end in itself.  It is likely that some topological / dynamical system / geometric information is desirable, and thus can be used to either understand current optimization methods or heuristically observed improvements.  For example, rapidly fluctuating topology durring optimization is an indicator of instability and need for regularization.  In general, perhaps simple topology is preferable for robustness.  Our observation that data flowing under the gradient system is attract to regions far from the original data as shown in Figure~\ref{digitIterations} may inspire new methods.  It appears that there the attracting set has a strongly attracting direction/manifold that is far from the data; data moves quickly to this manifold and then follows this manifold slowly to the attracting set.  It may be preferable to have this attracting manifold approximate the original data, and it may be that the strong manifold is connected to the weights - for example, a submanifold where some of the neuron outputs are zero.  Perhaps these are useful for deriving improved optimization methods or give improved understanding to currently observed useful methods.

\bibliography{my_refs3}

\end{document}